\documentclass[runningheads]{llncs}

\usepackage{eccv}

\usepackage{eccvabbrv}

\usepackage{graphicx}
\usepackage{booktabs}
\usepackage{booktabs}
\usepackage{multirow}
\usepackage{wrapfig,float,lipsum,booktabs}
\usepackage{lipsum}
\usepackage{amssymb}
\usepackage{graphicx}
\usepackage{subcaption}% <-- added

\usepackage[accsupp]{axessibility}  % Improves PDF readability for those with disabilities.

\usepackage{hyperref}

\usepackage{orcidlink}

\begin{document}

\title{One-stage Prompt-based Continual Learning} 

\titlerunning{One-stage Prompt-based Continual Learning}

\author{Youngeun Kim \and
Yuhang Li \and
Priyadarshini Panda}

\authorrunning{Kim et al.}

\institute{Yale University, CT, USA \\
\email{\{youngeun.kim,yuhang.li,priya.panda\}@yale.edu}}

\maketitle

\begin{abstract}
  Prompt-based Continual Learning (PCL) has gained considerable attention as a promising continual learning solution as it achieves state-of-the-art performance while preventing privacy violation and memory overhead issues. Nonetheless, existing PCL approaches face significant computational burdens because of two Vision Transformer (ViT) feed-forward stages; one is for the query ViT that generates a prompt query to select prompts inside a prompt pool; the other one is a backbone ViT that mixes information between selected prompts and image tokens. To address this, we introduce a one-stage PCL framework by directly using the intermediate layer's token embedding as a prompt query. This design removes the need for an additional feed-forward stage for query
ViT, resulting in $\sim 50\%$ computational cost reduction for both training and inference with marginal accuracy drop ($\le 1\%$).
We further introduce a Query-Pool Regularization (QR) loss that regulates the relationship between the prompt query and the prompt pool to improve representation power. The QR loss is only applied during training time, so there is no computational overhead at inference from the QR loss.
With the QR loss, our approach maintains $\sim 50\%$ computational cost reduction during inference as well as outperforms the prior two-stage PCL methods by $\sim 1.4\%$ on public class-incremental continual learning benchmarks including CIFAR-100, ImageNet-R, and DomainNet.
\end{abstract}

\section{Introduction}
Training models effectively and efficiently on a continuous stream of data presents a significant practical hurdle. A straightforward approach would entail accumulating both prior and new data and then updating the model using this comprehensive dataset. However, as data volume grows, fully retraining a model on such extensive data becomes increasingly impractical \cite{mai2022online,hadsell2020embracing}. Additionally, storing past data can raise privacy issues, such as those highlighted by the EU General Data Protection Regulation (GDPR) \cite{voigt2017eu}. An alternative solution is to adapt the model based solely on currently incoming data, eschewing any access to past data. This paradigm is termed as \textit{rehearsal-free continual learning} \cite{choi2021dual,gao2022r,yin2020dreaming,smith2021always,wang2021learning,wang2022dualprompt,smith2023coda}, and the primary goal is to diminish the effects of catastrophic forgetting on previously acquired data.

Among the rehearsal-free continual learning methods, Prompt-based Continual Learning (PCL) stands out as it has demonstrated state-of-the-art performance in image classification tasks, even surpassing rehearsal-based methods \cite{wang2021learning,wang2022dualprompt,smith2023coda,wang2022s}. PCL utilizes a pre-trained Vision Transformer (ViT) \cite{dosovitskiy2020image} and refines the model by training learnable tokens on the given data. PCL adopts a prompt pool-based training scheme where different prompts are selected and trained for each continual learning stage.
% a partial training scheme for the prompts within a prompt pool and freeze them to retain the information of the stage. 
This strategy enables a model to learn the information of the training data in a sequential manner with less memory overhead, as the prompt pool requires minimal resources.

\begin{figure}[t]
\begin{center}
\centering
\def\arraystretch{0.5}
\begin{tabular}{@{}c}
% \hspace{-4mm}
\includegraphics[width=0.68\linewidth]{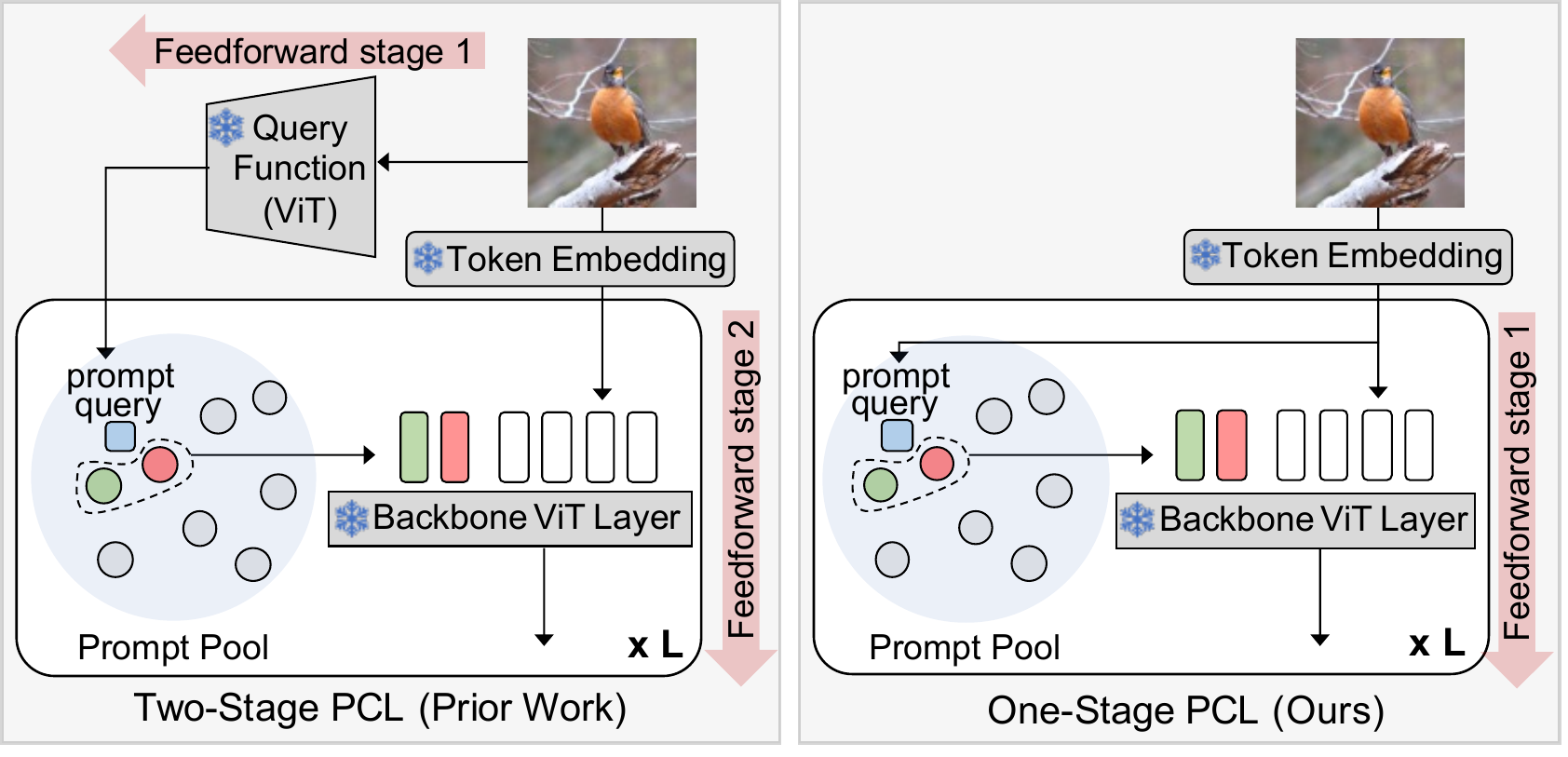}
\includegraphics[width=0.30\linewidth]{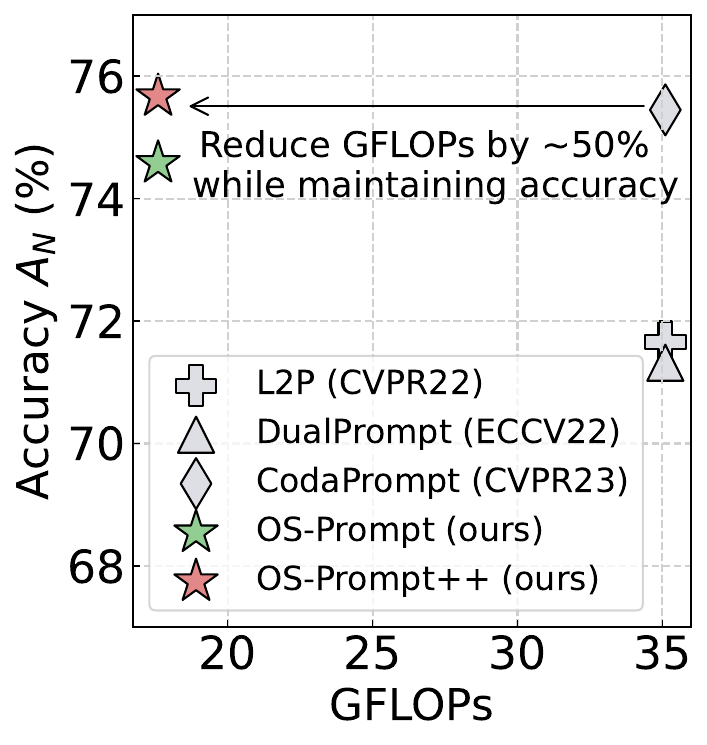} 
\\
\end{tabular}
\end{center}
\vspace{-4mm}
\caption{ 
Difference between prior PCL work and ours. Prior PCL work (\textbf{Left}) has two feed-forward stages for (1) a query function (ViT) to select input-specific prompts and (2) a backbone ViT layer to perform prompt learning with the selected prompts and image tokens. 
On the other hand, our one-stage PCL framework (\textbf{Middle}) uses an intermediate layer's token embedding as a prompt query so that it requires only one backbone ViT feed-forward stage.
As a result, our method reduces GFLOPs by $\sim 50\%$ compared to prior work while maintaining accuracy (\textbf{Right}).
}
 \vspace{-2mm}
\label{fig:concept_difference}
\end{figure}

Although PCL methods show state-of-the-art performance, huge computational costs from the two ViT feed-forward stages make the model difficult to deploy into resource-constrained devices \cite{harun2023efficient,wang2022sparcl,pellegrini2021continual}. 
Specifically, the PCL method requires two-stage ViT feedforward steps. One is for the query function that generates a prompt query. The other one is a backbone ViT that mixes information between selected prompts and input image tokens. We refer to this approach as a \textit{two-stage PCL} method, illustrated in Fig. \ref{fig:concept_difference} {Left}.  
% Making continual learning frameworks efficient is essential because real-world applications often demand rapid adaptability on IoT devices.

To address this limitation, we propose a \textit{one-stage PCL} framework with only one ViT feed-forward stage (Fig. \ref{fig:concept_difference} {Middle}), called OS-Prompt. Rather than deploying a separate ViT feed-forward phase to generate a prompt query, we directly use the intermediate layer's token embedding as a query. This is based on our observation that early layers show marginal shifts in the feature space during continual prompt learning (details are in Section \ref{sec:dist_exp}). 
% In other words, although we prepend prompts in the early layer, we still can obtain similar query features in the early layer across continual learning stages.
This observation enables us to use the intermediate token embedding as a prompt query which requires consistent representation across continual learning to minimize catastrophic forgetting.
% For example, the query (layer feature) on task $N$ is different from task $N-1$ as the prompts are updated during training.
% Such consistent feature representation minimizes catastrophic forgetting across continual learning , allowing us to use it as a query. 
Surprisingly, OS-Prompt shows a marginal performance drop (less than 1\%) while saving $\sim 50\%$ of computational cost (Fig. \ref{fig:concept_difference} Right).

As our OS-Prompt uses the intermediate layer's token embedding as a query instead of the last layer's token embedding (which is used in prior PCL methods), there is a slight performance drop due to the lack of representation power. To address this, we introduce a Query-Pool Regularization (QR) loss, which is designed to enhance the representation power of a prompt pool by utilizing the token embedding from the last layer. Importantly, the QR loss is applied only during training, ensuring no added computational burden during inference. We refer to our enhanced one-stage framework with QR loss as OS-Prompt++. Our OS-Prompt++ bridges the accuracy gap, performing better than OS-Prompt.

% Furthermore, the QR loss opens up the possibility of using a stronger ViT query function without increasing computational cost at inference.
% We use the advanced ViT models including Swin-Transformer \cite{liu2021swin}, PoolFormer \cite{yu2022metaformer} or ViT models trained on large-scale data \cite{radford2021learning}.
% By leveraging these advanced ViT models, we further improve the performance of the one-stage PCL framework. We highlight that the QR loss is architecture-agnostic with respect to the query function; it only needs a feed-forward operation to generate a query feature. This inherent flexibility is advantageous, as it permits seamless transitions to upcoming ViT models without the need for substantial architectural modification.

Overall, our contribution can be summarized as follows:
(1) We raise a drawback of the current PCL methods — high computational cost, particularly due to the two distinct ViT feed-forward stages. 
 As a solution to the computational inefficiency in existing PCL techniques, we propose OS-Prompt that reduces the computational cost by nearly 50\% without any significant performance drop. 
(2) To counter the slight performance degradation observed in our one-stage PCL framework, we introduce a QR loss. This ensures that the prompt pool maintains similarity with token embedding from both intermediate and final layers. Notably, our QR loss avoids any additional computational overhead during inference.
(3) We conduct experiments with rehearsal-free continual learning setting on CIFAR-100 \cite{krizhevsky2009learning}, ImageNet-R \cite{hendrycks2021many}, and DomainNet \cite{peng2019moment} benchmarks, outperforming the previous SOTA method CodaPrompt \cite{smith2023coda} by  $\sim 1.4\%$ with $\sim 50\%$ computational cost saving.

\section{Related Work}

\subsection{Continual Learning}
\vspace{-1mm}
 For a decade, continual learning has been explored as an important research topic, concentrating on the progressive adaptation of models over successive tasks or datasets.
One of the representative methods to address continual learning is the regularization-based method \cite{aljundi2018memory,li2017learning,zenke2017continual}. By adding a regularization loss between the current model and the previous model, 
these methods aim to minimize catastrophic forgetting. However, their performance is relatively low on challenging datasets compared to the other continual learning methods.
An alternative approach proposes to expand the network architecture as it progresses through different continual learning stages \cite{li2019learn,rusu2016progressive,yoon2017lifelong,serra2018overcoming}. Though these typically surpass the results of regularization-based methods, they come at the expense of significant memory consumption due to the increased parameters.
Rehearsal-based continual learning has introduced a component for archiving previous data \cite{chaudhry2018efficient,chaudhry2019tiny,hayes2019memory,buzzega2020dark,rebuffi2017icarl}. By leveraging accumulated data during the subsequent stages, they often outperform other continual learning methods. However, saving images causes memory overhead and privacy problems.
Given the inherent limitations, a growing interest is observed in rehearsal-free continual learning. These are crafted to address catastrophic forgetting without access to past data. Most of them propose methods to generate rehearsal images \cite{choi2021dual,gao2022r,yin2020dreaming,smith2021always}, but generating images is resource-heavy and time-consuming.
Recently, within the rehearsal-free domain, Prompt-based Continual Learning (PCL) methods have gained considerable attention because of their performance and memory efficiency.
By utilizing a pre-trained ViT model, they train only a small set of parameters called prompts to maintain the information against catastrophic forgetting.
For example, L2P \cite{wang2021learning} utilizes a prompt pool, selecting the appropriate prompts for the given image. Extending this principle, DualPrompt \cite{wang2022dualprompt} proposes general prompts for encompassing knowledge across multiple continual learning stages. Advancing this domain further, \cite{smith2023coda} facilitates end-to-end training of the prompt pool, achieving state-of-the-art performance. However, huge computational costs from the two ViT feed-forward stages make the model difficult to deploy into resource-constrained devices. Our work aims to resolve such computational complexity overhead in PCL.

\subsection{Prompt-based Learning}
\vspace{-1mm}

The efficient fine-tuning method for large pre-trained models has shown their practical benefits across various machine learning tasks \cite{rebuffi2018efficient, zhang2020side, zhang2021tip, zhou2022learning, he2022parameter,hu2021lora}. Instead of adjusting all parameters within neural architectures, the emphasis has shifted to leveraging a minimal set of weights for optimal transfer outcomes. In alignment with this, multiple methodologies \cite{rusu2016progressive, cai2020tinytl} have integrated a streamlined bottleneck component within the transformer framework, thus constraining gradient evaluations to select parameters. Strategies like TinyTL \cite{cai2020tinytl} and BitFit \cite{zaken2021bitfit} advocate for bias modifications in the fine-tuning process.
% Conversely, some techniques \cite{zhang2020side, sung2022lst} introduce auxiliary networks, designed for optimization, without altering the primary expansive model.
In a recent shift, prompt-based learning 
\cite{jia2022visual,khattak2023maple} captures task-specific knowledge with much smaller additional parameters than the other fine-tuning methods \cite{wang2021learning}.
Also, prompt-based learning only requires storing several prompt tokens, which are easy to plug-and-play, and hence, they are used to construct a prompt pool for recent continual learning methods.

\section{Preliminary}

\subsection{Problem Setting} 
\vspace{-1mm}
In a continual learning setting, a model is trained on ${T}$ continual learning stages with dataset $D = \{D_1, D_2, ..., D_T \}$, where $D_{t}$ is the data provided in $t$-th stage. Our problem is based on an image recognition task, so each data $D_{t}$ consists of pairs of images and labels. Also, data from the previous tasks is not accessible for future tasks.
Following the previous rehearsal-free continual learning settings  \cite{wang2022dualprompt,smith2023coda}, we focus on the class-incremental continual learning setting where task identity is
unknown at inference. This is a challenging scenario compared to others such as task-incremental continual learning where the task labels are provided for both training and test phases \cite{hsu2018re}.
For the experiments, we split classes into $T$ chunks with continual learning benchmarks including CIFAR-100 \cite{krizhevsky2009learning} and ImageNet-R \cite{hendrycks2021many}.

\subsection{Two-stage Prompt-based Continual Learning}

In our framework, we focus on improving efficiency through a new query selection rather than changing the way prompts are formed from the prompt pool with the given query, a common focus in earlier PCL studies \cite{wang2022dualprompt,wang2021learning,smith2023coda}.
For a fair comparison with earlier work, we follow the overall PCL framework established in previous studies.
Given that our contribution complements the prompt formation technique, our method can seamlessly work with future works that propose a stronger prompt generation method.

The PCL framework selects the input-aware prompts from the layer-wise prompt pool and adds them to the backbone ViT to consider task knowledge. The $l$-th layer has the prompt pool $P_l=\{k^1_l:p^1_l, ...,k^M_l: p^M_l\}$ which contains $M$ prompt components $p \in \mathbb{R}^{L_p \times D}$ and the corresponding key $k \in \mathbb{R}^{D}$. Here, $L_p$ and $D$ stand for the prompt length and the feature dimension, respectively.

The prior PCL method consists of two feed-forward stages. In the first stage, a prompt query $q\in \mathbb{R}^{D}$ is extracted from pre-trained query ViT $Q(\cdot)$, utilizing the $[CLS]$ token from the final layer.
\begin{equation}
    q = Q(x)_{[CLS]}.
    \label{eq:get_query}
\end{equation}
Here $x$ is the given RGB image input. The extracted prompt query is used to form a prompt $\phi_l \in \mathbb{R}^{L_p \times D}$ 
 from a prompt pool $P_l$, which can be formulated as:
\begin{equation}
    \phi_l = g(q, P_l).
\end{equation}
The prompt formation function $g(\cdot)$ has been a major contribution in the previous literature.
L2P \cite{wang2021learning} and DualPrompt \cite{wang2022dualprompt} select prompts having top-N similarity (\eg cosine similarity) between query $q$ and prompt key $k_l$. 
The recent state-of-the-art CodaPrompt \cite{smith2023coda} conducts a weighted summation of prompt components based on their similarity to enable end-to-end training.

The obtained prompt $\phi_l$ for layer $l$ is given to backbone ViT layer $f_l$ with input token embedding $x_l$. This is the second feed-forward stage of the prior PCL methods.
\begin{equation}
    x_{l+1} = f_l(x_l, \phi_l).
    \label{eq:compute_backbone}
\end{equation}
Following DualPrompt \cite{wang2022dualprompt} and CodaPrompt \cite{smith2023coda}, we use prefix-tuning to add the information of a prompt $\phi_l$ inside $f_l$.
The prefix-tuning splits prompt $\phi_l$ into  $[\phi_k, \phi_v] \in \mathbb{R}^{\frac{L_p}{2} \times D}$, and then prepends them to the key and value inside the self-attention block of ViT. 
We utilize Multi-Head Self-Attention (MHSA) \cite{dosovitskiy2020image} like the prior PCL work, which computes the outputs from multiple single-head attention blocks.
\begin{equation}
    MHSA(x, \phi_k, \phi_v) = Concat[head_1, ..., head_H]W_o.
\end{equation}
\begin{equation}
    head_i = Attention(xW_q^i,[\phi_k;x]W_k^i,[\phi_v;x]W_v^i).
\end{equation}
Here, $W_o, W_q, W_k, W_v$ are the projection matrices. This prefix-tuning method is applied to 
 the first 5 layers of backbone ViT. The leftover layers conduct a standard MHSA without prompts.

However, the prior PCL approach requires two-stage ViT feedforward (Eq. \ref{eq:get_query} and Eq. \ref{eq:compute_backbone}), which doubles computational cost. In our work, we aim to improve the computational efficiency of the PCL framework without degrading the performance.

\begin{figure}[t]
\begin{center}
\centering
\def\arraystretch{0.5}
\begin{tabular}{@{}c}
% \hspace{-4mm}
\includegraphics[width=0.99\linewidth]{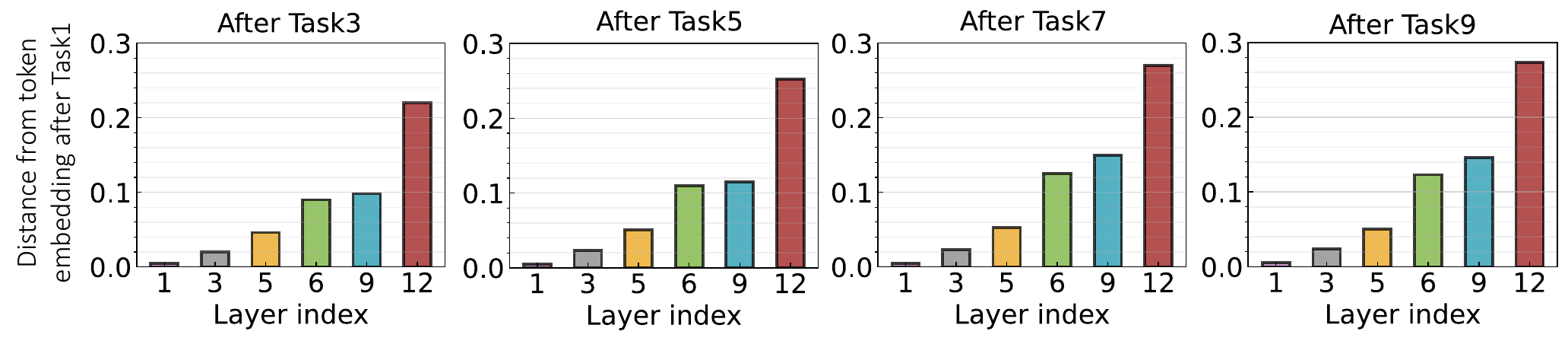} 
\\
\end{tabular}
\end{center}
\vspace{-4mm}
\caption{ 
We measure layer-wise feature distances between token embeddings of the model after training on task 1 and when a new task is learned. Each column represents the embeddings after training a model on Tasks 3, 5, 7, and 9. For instance, the left figure represents the distance between the token embeddings of Task 1 and those learned when Task 3 is completed.
We train prompts using CodaPrompt \cite{smith2023coda} and use $1-CosSim(x, y)$ to measure the layer-wise distance on the training dataset. We use a CIFAR100 10-task setting.
{We provide more examples in the Supplementary Materials.}
% We illustrate how the token embedding deviates from its initial value throughout the continual learning process. Each task has 20 or 50 training epochs, depending on the dataset. 
% We train prompts using CodaPrompt \cite{smith2023coda} and use $1-CosSim(x, y)$ to measure the layer-wise distance on the training dataset.
}
 \vspace{-3mm}
\label{fig:feature_dist_across_CL}
\end{figure}

\section{One-stage Prompt-based Continual Learning}

We propose to reduce the computational cost of PCL by restructuring two-step feedforward stages. To this end, we propose a new one-stage PCL framework called \textit{OS-Prompt}. Instead of using a separate feedforward step to compute the prompt query $q$ from Eq. \ref{eq:get_query}, we take a token embedding from the intermediate layer of a backbone ViT as the query (illustrated in Fig. \ref{fig:concept_difference}).

\subsection{How Stable are Token Embeddings across Continual Learning?}
\label{sec:dist_exp}

We first ensure the validity of using intermediate token embedding as a prompt query.
The original two-stage design employs a frozen pre-trained ViT, ensuring consistent prompt query representation throughout continual learning.
It is essential to maintain a consistent (or similar) prompt query representation because changes in prompt query would bring catastrophic forgetting.
In our approach, token embeddings in the backbone ViT continually change as prompt tokens are updated during learning. For instance, after training on tasks $1$ and $2$, prompts have different values, resulting in varying token embeddings for an identical image. Consequently, it is crucial to assess if token embeddings sustain a consistent representation throughout the continual learning.

To address the concern, we validate the deviation of the token embedding as prompt continual learning progresses.  In Fig. \ref{fig:feature_dist_across_CL}, we show how this change happens at each layer during continuous learning tasks.
From our study, two main observations can be highlighted:
(1) When we add prompts, there is a larger difference in the deeper layers. For example, layers $1 \sim 5$ have a small difference ($\le$ 0.1). However, the last layer shows a bigger change ($\ge$ 0.1).
(2) As learning continues, the earlier layers remain relatively stable, but the deeper layers change more.
This observation concludes that although prompt tokens are included, the token embedding of early layers shows minor changes during training. Therefore, using token embeddings from these earlier layers would give a stable and consistent representation throughout the continual learning stages.

\subsection{One-stage PCL Framework}

\begin{figure}[t]
\begin{center}
\centering
\def\arraystretch{0.5}
\begin{tabular}{@{}c}
% \hspace{-4mm}
\includegraphics[width=0.9\linewidth]{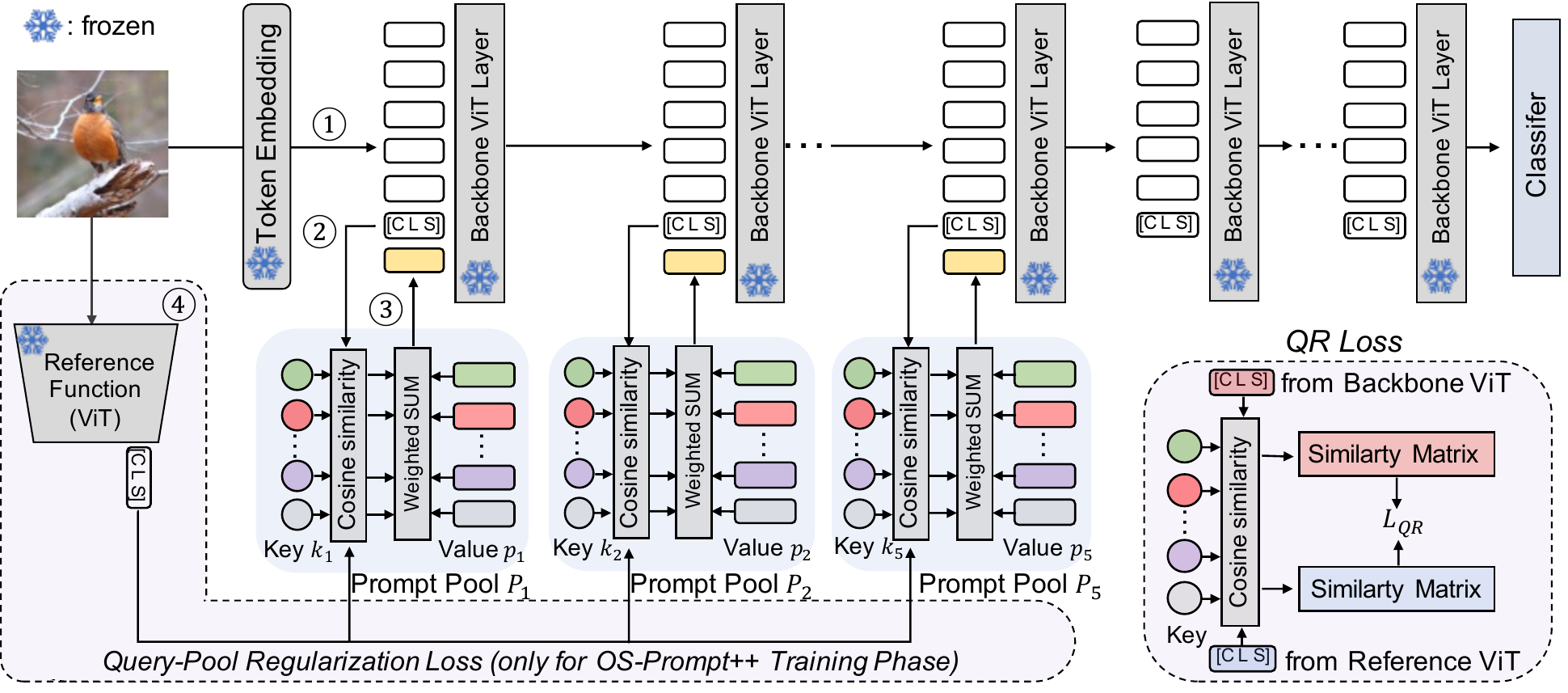} 
\\
\end{tabular}
\end{center}
\vspace{-3mm}
\caption{ 
Our OS-Prompt (OS-Prompt++) framework. An image passes through the backbone ViT layers to get the final prediction.
From layer 1 to layer 5, we prepend prompt tokens to the image tokens, which can be obtained in the following progress:
\raisebox{.5pt}{\textcircled{\raisebox{-.9pt} {1}}}  We first compute the image token embedding from the previous layer.
\raisebox{.5pt}{\textcircled{\raisebox{-.9pt} {2}}} We use $[CLS]$ token as a prompt query used for measuring cosine similarity between prompt keys inside the prompt pool.
\raisebox{.5pt}{\textcircled{\raisebox{-.9pt} {3}}} Based on the similarity, we do weighted sum values to obtain prompt tokens.
\raisebox{.5pt}{\textcircled{\raisebox{-.9pt} {4}}}
To further improve the accuracy, we present OS-Prompt++ .
We integrate the query-pool regularization loss (dotted line), enabling the prompt pool to capture a stronger representation from the reference ViT.
}
 \vspace{-1mm}
\label{fig:method:overall}
\end{figure}

Building on these observations, we employ the token embedding from the early layers as a prompt query to generate layer-wise prompt tokens. For a fair comparison, we implement prompts across layers 1 to 5, in line with prior work.
The proposed OS-Prompt framework is illustrated in Fig. \ref{fig:method:overall}. For each layer, given the input token embedding, we directly use the $[CLS]$ token as the prompt query. The original  query selection equation (Eq. \ref{eq:get_query}) can be reformulated as:
\begin{equation}
    q_l = x_{l_{[CLS]}}.
    \label{eq:get_query_ours}
\end{equation}
Using the provided query $q_l$, we generate a prompt from the prompt pool following the state-of-the-art CodaPrompt \cite{smith2023coda}. It is worth highlighting that our primary contribution lies not in introducing a new prompt generation technique but in presenting a more efficient framework for query selection.
We first measure the cosine similarity $\gamma(\cdot)$ between $q_l$ and keys $\{k^1_l, ..., k^M_l\}$, and then we perform a weighted summation of the corresponding value $p^m_l$ (\ie prompt) based on the similarity. 
\begin{equation}
    \phi_l = \sum_{m} \gamma(q_l, k^m_l) p^m_l.
\end{equation}
The generated prompt is then prepended to the image tokens. Notably, since we produce a prompt query without the need for an extra ViT feedforward, the computational overhead is reduced by approximately $50\%$ for both training and inference. % with only a marginal performance drop of $\le 1\%$.

\subsection{Query-Pool Regularization Loss}

Our OS-Prompt relies on intermediate token embeddings. As a result, the prompt query exhibits diminished representational capacity compared to previous PCL approaches that utilize the $[CLS]$ token from the final layer. This reduced capacity for representation brings performance degradation.
To mitigate this, we introduce a Query-Pool Regularization (QR) loss. The QR loss ensures that the query-pool relationship becomes similar to that of the final layer's $[CLS]$ token, thereby improving representation power. 
We extract the $[CLS]$ token from the last layer of a reference ViT architecture $R(\cdot)$ and employ it as our reference prompt query $r \in \mathbb{R}^{1 \times D}$ where $D$ is the feature dimension:
\begin{equation}
    r = R(x).
    \label{eq:get_reference}
\end{equation}
Let $K_l \in \mathbb{R}^{M \times D}$ be the matrix representation of prompt keys at layer $l$. We then define a similarity matrix $A^l_{query} \in \mathbb{R}^{M \times 1}$ to capture the relationship between query $q_l\in \mathbb{R}^{1 \times D}$ (from Eq. \ref{eq:get_query_ours}) and prompt keys. Similarly, we compute the similarity matrix $A^l_{ref}$ to represent the relationship between the reference prompt query $r$ and the prompt keys.
\begin{equation}
    A^l_{query} = Softmax(\frac{K_lq_l^T}{||K_l||_2||q_l||_2})   
    \hspace{7mm} A^l_{ref} = Softmax(\frac{K_lr^T}{||K_l||_2||r||_2}).
\end{equation}
We apply $Softmax()$ to measure the relative similarity among query-key pairs, considering that $q$ and $r$ show distinct feature distributions originating from different layers.
The QR loss penalizes the distance between two similarity matrices.
\begin{equation}
    \mathcal{L}_{QR} = \sum_l ||A^l_{query} - A^l_{ref}||^2_2.
    \label{eq:qrloss}
\end{equation}  
The QR loss is added to the cross-entropy loss $\mathcal{L}_{CE}$ for classification.
The total loss function can be written as:
\begin{equation}
    \mathcal{L}_{total} = \mathcal{L}_{CE} + \lambda \mathcal{L}_{QR},
    \label{eq:total_loss}
\end{equation}
where $\lambda$ is a hyperparameter to balance between two loss terms.
Importantly, the QR loss is applied during the training phase, ensuring that there is no computational overhead during inference. Note that we only train prompts within a prompt pool while freezing the other weight parameters.

\begin{table}[t]
   \centering
% \small
 \aboverulesep=0.1ex % Solution part 1 of 3
   \belowrulesep=0ex % Solution part 1 of 3
   \caption{Results on ImageNet-R for different task configurations: 5 tasks (40 classes/task), 10 tasks (20 classes/task), and 20 tasks (10 classes/task). $A_N$ represents the average accuracy across tasks, and $F_N$ indicates the mean forgetting rate. $\uparrow$ and $\downarrow$ indicate whether a metric is better with a higher or lower value, respectively. We average values over five runs with different seeds.
}
\vspace{-0mm}
\resizebox{0.88\textwidth}{!}{%
\begin{tabular}{l|cc|cc|cc}

% \hlinewd{1pt}
 \toprule
Setting &  \multicolumn{2}{c|}{Task-5} & \multicolumn{2}{c|}{Task-10} &\multicolumn{2}{c}{Task-20} \\
 \midrule
Method & $A_N(\uparrow)$ & $F_N(\downarrow)$ &  $A_N(\uparrow)$ & $F_N(\downarrow)$  &  $A_N(\uparrow)$ & $F_N(\downarrow)$  \\
 \midrule
UB & 77.13 & - & 77.13 & - & 77.13 & -\\
\midrule
FT &18.74 ± 0.44 & 41.49 ± 0.52  & 10.12 ± 0.51 &25.69 ± 0.23 & 4.75 ± 0.40 & 16.34 ± 0.19 \\
ER & 71.72 ± 0.71 & 13.70 ± 0.26 & 64.43 ± 1.16  & 10.30 ± 0.05  & 52.43 ± 0.87 & 7.70 ± 0.13\\
LwF & 74.56 ± 0.59 & 4.98 ± 0.37 & 66.73 ± 1.25 & 3.52 ± 0.39 & 54.05 ± 2.66 & 2.86 ± 0.26\\
L2P & 70.83 ± 0.58 & 3.36 ± 0.18 &69.29 ± 0.73 & 2.03 ± 0.19 & 65.89 ± 1.30&  1.24 ± 0.14\\
Deep L2P &73.93 ± 0.37& 2.69 ± 0.10 &71.66 ± 0.64 &1.78 ± 0.16 &68.42 ± 1.20 &1.12 ± 0.13  \\
DualPrompt &73.05 ± 0.50& 2.64 ± 0.17 &71.32 ± 0.62& 1.71 ± 0.24& 67.87 ± 1.39& 1.07 ± 0.14\\
CodaPrompt & 76.51 ± 0.38& 2.99 ± 0.19& 75.45 ± 0.56&1.64 ± 0.10 &72.37 ± 1.19& 0.96 ± 0.15 \\
 \midrule
OS-Prompt & 75.74 ± 0.58 &3.32 ± 0.31 & 74.58 ± 0.56& 1.92 ± 0.15& 72.00 ± 0.60&
1.09 ± 0.11\\
OS-Prompt++ & \textbf{77.07 ± 0.15} & \textbf{2.23 ± 0.18} & \textbf{75.67 ± 0.40} & \textbf{1.27 ± 0.10} & \textbf{73.77 ± 0.19} & \textbf{0.79 ± 0.07}\\
% \hlinewd{1pt}
 \bottomrule
\end{tabular}%
}
\label{table:exp:results_imgnetr}
  \vspace{-3mm}
\end{table}

\begin{table}[t]
 \aboverulesep=0.1ex % Solution part 1 of 3
   \belowrulesep=0ex % Solution part 1 of 3
\begin{minipage}[b]{0.40\linewidth}\centering
   \caption{Accuracy comparison on CIFAR-100 10-task setting.
}
\vspace{-1.5mm}
\resizebox{0.9\textwidth}{!}{
\begin{tabular}{l|cc}
% \hlinewd{1pt}
 \toprule
% Setting &  \multicolumn{2}{c}{Task-10} \\
%  \midrule
Method  & \hspace{3mm} $A_N(\uparrow)$ \hspace{3mm}& 
 \hspace{3mm} $F_N(\downarrow)$ \hspace{3mm}  \\
 \midrule
UB & 89.30 & - \\
 \midrule
ER & 76.20 ± 1.04 & 8.50 ± 0.37 \\
% FT & 9.92 ± 0.27 & 29.21 ± 0.18  \\
% LwF & 64.83 ± 1.03  & 5.27 ± 0.39 \\
Deep L2P & 84.30 ± 1.03& 1.53 ± 0.40\\
DualPrompt & 83.05 ± 1.16& 1.72 ± 0.40\\
CodaPrompt & 86.25 ± 0.74& 1.67 ± 0.26 \\
 \midrule
OS-Prompt &  86.42 ± 0.61 &1.64 ± 0.14 \\
{OS-Prompt++} & \textbf{86.68 ± 0.67} & \textbf{1.18 ± 0.21} \\
% \hlinewd{1pt}
 \bottomrule
\end{tabular}%
 \label{table:exp:results_cifar}
}
\end{minipage}
\hspace{0.05cm}
\begin{minipage}[b]{0.562\linewidth}
\caption{ GFLOPs comparison of PCL works. We provide relative cost (\%) with respect to L2P.
}
\vspace{-1.5mm}
\centering
\resizebox{0.91\textwidth}{!}{
\begin{tabular}{l|cc}
% \hlinewd{1pt}
 \toprule
{Method} &Training GFLOPs  & Inference  GFLOPs \\
% \multirow{ 2}{*}{Method} & Training   & Inference  &Training  & Inference   \\
%  &   Latency &  Latency & GFLOPs &  GFLOPs  \\
 % &   (ms) &  (ms) &  &    \\
 \midrule
L2P &  {52.8} (100\%) &35.1 (100\%)\\
Deep L2P & {52.8} (100\%)&35.1 (100\%) \\
DualPrompt &{52.8} (100\%) &35.1 (100\%) \\
CodaPrompt& {52.8} (100\%) &35.1 (100\%)\\
 \midrule
{OS-Prompt} &\textbf{{35.4} (66.7\%)} &\textbf{17.6  (50.1\%)} \\
{OS-Prompt++}  &{{52.8} (100\%)}&\textbf{17.6  (50.1\%)}\\
% \hlinewd{1pt}
 \bottomrule
\end{tabular}
 \label{table:exp:efficiency}
}
\end{minipage}
  \vspace{-2mm}
\end{table}

\section{Experiments}
\vspace{-2mm}
\subsection{Experiment Setting}
\vspace{-1mm}

\noindent\textbf{Dataset.}
We utilize the Split CIFAR-100 \cite{krizhevsky2009learning} and Split ImageNet-R \cite{hendrycks2020many} benchmarks for class-incremental continual learning. The Split CIFAR-100 partitions the original CIFAR-100 dataset into 10 distinct tasks, each comprising 10 classes. The Split ImageNet-R benchmark is an adaptation of ImageNet-R, encompassing diverse styles, including cartoon, graffiti, and origami. For our experiments, we segment ImageNet-R into 5, 10, or 20 distinct class groupings. Given its substantial intra-class diversity, the ImageNet-R benchmark is viewed as a particularly challenging benchmark.
We also provide experiments on DomainNet \cite{peng2019moment}, a large-scale domain incremental learning dataset. The dataset consists of around 0.6 million images distributed across 345 classes. We use 5 different domains (Clipart → Real → Infograph → Sketch → Painting) as a continual learning task where each task consists of 69 classes.

\noindent\textbf{Experimental Details.}
We performed our experiments using the ViT-B/16 \cite{dosovitskiy2020image} pre-trained on ImageNet-1k, a standard backbone in earlier PCL research. To ensure a fair comparison with prior work, we maintain the same prompt length (8) and number of prompt components (100) as used in CodaPrompt. Like CodaPrompt, we divide the total prompt components into the number of tasks, and provide the partial component for each task. In task $N$, the prompt components from task $1\sim N$ is frozen, and we only train the key and prompt components from task $N$. We split 20\% of the training set for hyperparameter $\lambda$ tuning in Eq. \ref{eq:total_loss}.
Our experimental setup is based on the PyTorch. The experiments utilize four Nvidia RTX2080ti GPUs.
For robustness, we conducted our benchmarks with five different permutations of task class order, presenting both the average and standard deviation of the results. {More detailed information can be found in the Supplementary Materials}.

\noindent\textbf{Evaluation Metrics.} We use two metrics for evaluation: (1) Average final accuracy $A_N$, which measures the overall accuracy across N tasks. \cite{wang2021learning,wang2022dualprompt,smith2023coda} (2) Average forgetting $F_N$, which tracks local performance drop over N tasks \cite{smith2023closer, smith2023coda,lopez2017gradient}. Note, $A_N$ is mainly used for performance comparison in literature.

\subsection{Comparison with prior PCL works}

We compare our OS-Prompt with prior continual learning methods. This includes non-PCL methods such as ER \cite{chaudhry2019tiny} and LWF \cite{li2017learning}, as well as PCL methods like L2P \cite{wang2021learning}, DualPrompt \cite{wang2022dualprompt}, and CodaPrompt \cite{smith2023coda}. We present both the upper bound (UB) performance and results from fine-tuning (FT). UB refers to training a model with standard supervised training using all data (no continual learning). FT indicates that a model undergoes sequential training with continual learning data without incorporating prompt learning.
Moreover, we report L2P/Deep L2P implementation from CodaPrompt \cite{smith2023coda}, both in its original form and improved version by applying prompt pool through layers 1 to 5. In our comparisons, \textit{OS-Prompt} represents our one-shot prompt framework without the QR loss, while \textit{OS-Prompt++} is with the QR loss.

\begin{wraptable}{h}{5.0cm}
\vspace{-11mm}
\caption{Performance comparison on
DomainNet 5-task setting.}
\resizebox{0.4\textwidth}{!}{
\begin{tabular}{l|cc}
% \hlinewd{1pt}
 \toprule
% Setting &  \multicolumn{2}{c}{Task-10} \\
%  \midrule
Method  & \hspace{3mm} $A_N(\uparrow)$ \hspace{3mm}& 
 \hspace{3mm} $F_N(\downarrow)$ \hspace{3mm}  \\
 \midrule
 UB & 79.65 & - \\
 \midrule
FT & 18.00 ± 0.26&
43.55 ± 0.27
 \\ 
 ER & 58.32 ± 0.47&
26.25 ± 0.24
 \\ 
%  L2P & 69.58 ± 0.39&
% 2.25 ± 0.08 \\
Deep L2P & 69.58 ± 0.39&
2.25 ± 0.08 \\
DualPrompt & 70.73 ± 0.49&
\textbf{2.03 ± 0.22} \\CodaPrompt & 73.24 ± 0.59&
3.46 ± 0.09  \\
 \midrule
OS-Prompt &  72.24 ± 0.13&
2.94 ± 0.02 
 \\
{OS-Prompt++} & \textbf{73.32 ± 0.32}&
2.07 ± 0.06
\\
% \hlinewd{1pt}
 \bottomrule
\end{tabular}%
}
 \label{table:exp:domainnet}
\vspace{-6mm}
\end{wraptable} %

Table \ref{table:exp:results_imgnetr} shows the results on ImageNet-R. Our OS-Prompt indicates only a slight performance drop ($\le 1\%$) across the 5-task, 10-task, and 20-task settings. The reason for the performance drop could be the reduced representational capacity of a prompt query from the intermediate token embedding. However, OS-Prompt++ effectively counters this limitation by incorporating the QR loss. Such an observation underscores the significance of the relationship between the query and the prompt pool in PCL.
OS-prompt++ shows slight performance improvement across all scenarios. This implies a prompt query that contains task information helps to enhance representation in the prompt pool, suggesting that exploring methods to integrate task information inside the prompt selection process could be an interesting research direction in PCL.
Table \ref{table:exp:results_cifar} and Table \ref{table:exp:domainnet} provide results on 10-task CIFAR-100 and DomainNet, respectively, which shows a similar trend with ImageNet-R. 
\begin{wrapfigure}{R}{0.45\textwidth}
     \vspace{-9mm}
     \centering
         \includegraphics[width=0.45\textwidth]{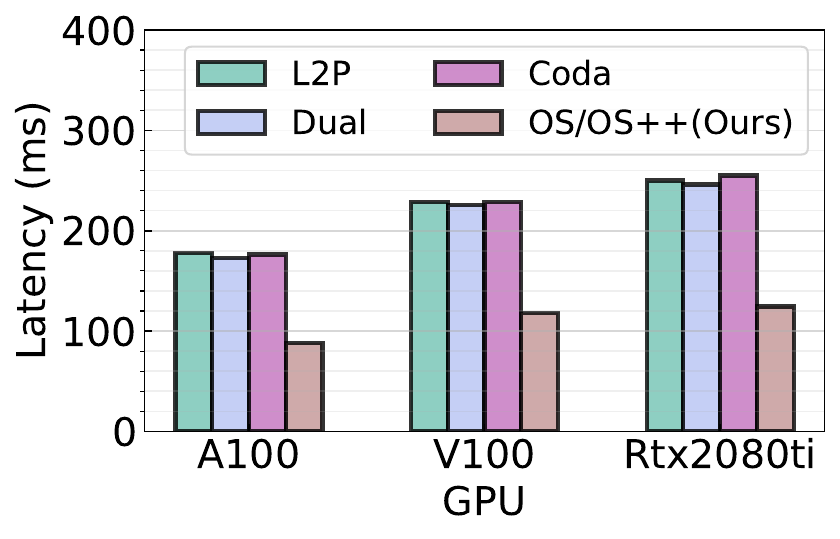}
        \vspace{-8mm}
 \caption{
Comparison of latency across PCL methods on different GPUs.
% , using a batch size of 32 on a single GPU.
 }
 \vspace{-6mm}
     \label{fig:exp:latency}
\end{wrapfigure}

In Table \ref{table:exp:efficiency}, we provide GFLOPs for various PCL methods, rounding GFLOPs to one decimal place. It is worth noting that while prior PCL methods might have marginally different GFLOP values, they are close enough to appear identical in the table. This implies that different prompt formation schemes proposed in prior works do not substantially impact GFLOPs. The table shows OS-Prompt operates at 66.7\% and 50.1\% of the GFLOPs, during training and inference, respectively, relative to prior methods. While OS-Prompt++ maintains a training GFLOPs count comparable to earlier work due to the added feedforward process in the reference ViT, it does not employ the reference ViT during inference, bringing it down to 50.1\% GFLOPs.
In Fig. \ref{fig:exp:latency}, we compare the inference GPU latency of previous PCL methods with our OS-Prompt across three distinct GPU setups. All measurements are taken using a batch size of 32. Similar to the trend observed in FLOPs, our approach reduces the latency by  $\sim 50\%$.

\subsection{Discussion on Training Computational Cost}

In this section,  we provide an in-depth discussion of the training computational cost of PCL methods. Following \cite{ghunaim2023real}, Table \ref{table:supply:training_cost} presents the relative training complexity of each method with respect to ER \cite{chaudhry2019tiny}, which is the simple baseline with standard gradient-based training.

\begin{table}[t]
 \aboverulesep=0.1ex % Solution part 1 of 3
   \belowrulesep=0ex % Solution part 1 of 3
\begin{minipage}[b]{0.52\linewidth}\centering
   \caption{Relative training computational cost with repect to ER.
}
\vspace{-1.5mm}
\resizebox{0.99\textwidth}{!}{
\begin{tabular}{l|c|c}
% \hlinewd{1pt}
 \toprule
% Setting &  \multicolumn{2}{c}{Task-10} \\
%  \midrule
\multirow{2}{*}{CL strategy}  & \multirow{2}{*}{Method} & Training \\
& & Complexity\\
 \midrule
Replay-based  & ER& 1
 \\
Regularization-based   & LWF & 4/3
\\
Prompt-based &L2P & 1
\\
Prompt-based  & DualPrompt & 1
\\ Prompt-based  & CodaPrompt & 1
\\
Prompt-based  &{OS-Prompt} (Ours) & 2/3
\\
Prompt-based  &{OS-Prompt++} (Ours)& 1
\\
% \hlinewd{1pt}
 \bottomrule
\end{tabular}%
\label{table:supply:training_cost}
}
\end{minipage}
\hspace{0.05cm}
\begin{minipage}[b]{0.48\linewidth}
\caption{ ImageNet-R 10-task results with unsupervised pre-trained model \cite{caron2021emerging}.}
\vspace{-1.5mm}
\centering
\resizebox{0.91\textwidth}{!}{
\begin{tabular}{l|cc}
% \hlinewd{1pt}
 \toprule
% Setting &  \multicolumn{2}{c}{Task-10} \\
%  \midrule
Method  & \hspace{3mm} $A_N(\uparrow)$ \hspace{3mm}& 
 \hspace{3mm} $F_N(\downarrow)$ \hspace{3mm}  \\
 \midrule
ER & 60.43 ± 1.16 & 13.30 ± 0.12 \\
LwF & 62.73 ± 1.13   & 4.32 ± 0.63 \\
L2P & 60.32 ± 0.56 &
2.30 ± 0.11\\
DualPrompt & 61.77 ± 0.61&2.31 ± 0.23
\\CodaPrompt & 67.61 ± 0.19&2.23 ± 0.29
 \\
 \midrule
OS-Prompt &  67.52 ± 0.34&2.32 ± 0.13
 \\
{OS-Prompt++} & 67.92 ± 0.42&2.19 ± 0.26
\\
% \hlinewd{1pt}
 \bottomrule
\end{tabular}%
 \label{table:supply:dino}
}
\end{minipage}
  \vspace{-2mm}
\end{table}

Here, we would like to clarify the training computational cost of prompt learning. In general neural network training,  the forward-backward computational cost ratio is approximately 1:2. This is due to gradient backpropagation ($\frac{dL}{da_l}=W_{l+1}\frac{dL}{da_{l+1}}$) and weight-updating ($\frac{dL}{dW_l}=\frac{dL}{da_{l+1}}a_l$), where $L$ represents the loss, $W$ denotes the weight parameter, $a$ is the activation, and $l$ is the layer index.  In prompt learning, the forward-backward computational cost ratio is approximately 1:1. This is in contrast to general neural network training, as only a small fraction of the weight parameters (less than 1\%) are updated.

Bearing this in mind, we present the observations derived from the table, assuming that all methods employ the same architecture.
(1) Previous PCL (L2P, DualPrompt, CodaPrompt) consists of two steps; First, the query ViT requires only feedforward without backpropagation; Second, the backbone ViT feedforward-backward training with prompt tuning. This results in the relative training computational cost is 1 ($= \frac{Query ViT_{fw} (1) + Backbone ViT_{fw} (1) + Backbone ViT_{bw} (1)}{ER_{fw} (1) + ER_{bw} (2)}$). Here, $fw$ and $bw$ denote forward and backward propagation respectively.
(2)
Similarly, our OS-Prompt++ has a reference ViT which only requires a feedforward step, bringing 1 relative training computational cost with respect to ER. 
(3) 
On the other hand, our OS-Prompt requires one feedforward-backward step like standard training. Therefore, relative training computaional cost becomes $\frac{2}{3}$ ($= \frac{Backbone ViT_{fw} (1) + Backbone ViT_{bw} (1)}{ER_{fw} (1) + ER_{bw} (2)}$). This shows the potential of our OS-Prompt on online continual learning.

{
Overall, we propose two versions of the one-stage PCL method (i.e., OS-prompt and OS-prompt++), and these two options enable the users can select a suitable method depending on the problem setting. For example, for online continual learning, OS-prompt is a better option since it requires less training cost. On the other hand, for offline continual learning, one can use OS-prompt++ to maximize the performance while spending more training energy. 
}

\subsection{Impact of Unsupervised Pre-trained Weights}
One underlying assumption of existing PCL methods is their reliance on supervised pretraining on ImageNet-1k. While this pre-training has a substantial impact on model performance, it may not always be feasible for a general continual learning task. One possible solution is using unsupervised pre-trained models.
To explore this, we use DINO pre-trained weights \cite{caron2021emerging} instead of ImageNet-1k pre-trained weights, and compare the performance across ER, LwF, and PCL works. We set the other experimental settings as identical.
We report the results with ImageNet-R 10-task in Table \ref{table:supply:dino}. Compared to the ImageNet-1k pre-trained model (with supervised training), performances across all methods degrade. This observation suggests that the backbone model plays a crucial role in PCL. Moreover, similar to the ImageNet-1k pre-trained model, CodaPrompt and our OS-Prompt outperform the other methods by a significant margin. This indicates that our method continues to perform well with various backbones trained on different methods.

\subsection{Analysis of Design Components}

\textbf{QR Loss Design.} To understand the effect of different components in our proposed QR loss (Eq. \ref{eq:qrloss}), we conducted an ablation study with different settings, summarized in Table \ref{table:exp:qr_design}. 
There are two main components in the QR loss: \textit{Cosine Similarity} and \textit{Softmax}, and we provide the accuracy of different four combinations.
Without Cosine Similarity and Softmax, our framework yields a performance of $75.00\%$ for $A_N$.
Adding Cosine Similarity or Softmax improves the performance, suggesting their collaborative role in enhancing the model's performance in our QR loss design. 
We also provide the sensitivity analysis of our method with respect to the hyperparameter $\lambda$ in Table \ref{table:exp:hyperparameterabl}. Across the three distinct task configurations on ImageNet-R, we note that the performance fluctuations are minimal, despite the variation in $\lambda$ values. These results underline the robustness of our method with respect to $\lambda$ value, suggesting that our approach is not sensitive to  hyperparameter. 

\begin{figure}[t]
 \aboverulesep=0.1ex % Solution part 1 of 3
   \belowrulesep=0ex % Solution part 1 of 3
\begin{minipage}[b]{0.64\linewidth}\centering
\vspace{-1mm}
\hspace{-2mm}
\centering
\includegraphics[width=0.43\textwidth]{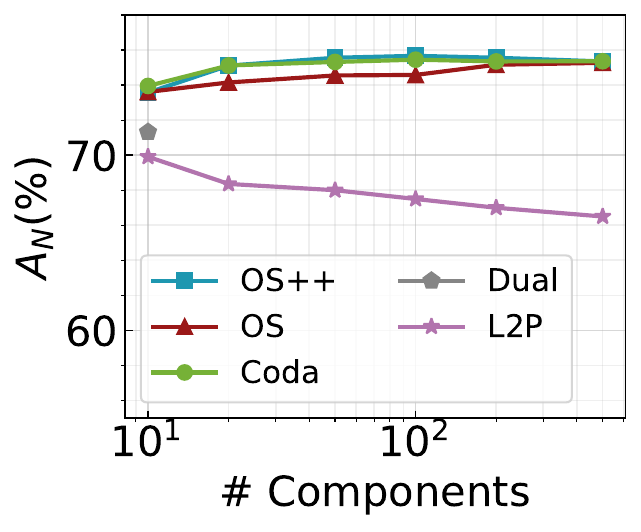}
\hspace{2mm}
\includegraphics[width=0.43\textwidth]{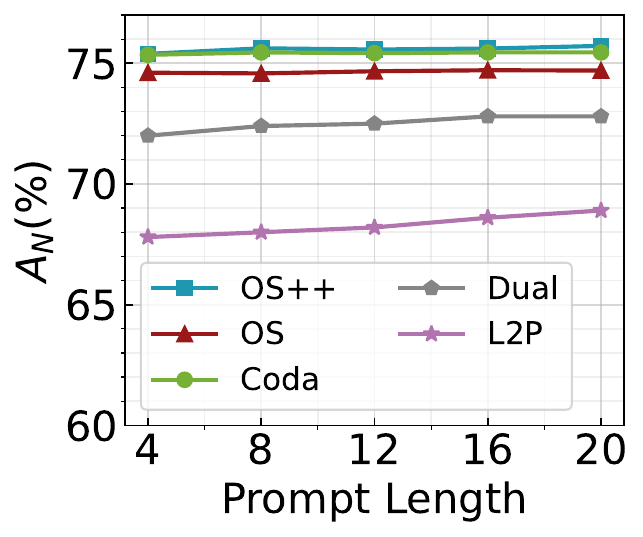}
        \vspace{-3mm}
 \caption{
 {
Analysis of accuracy $A_N$ with respect to prompt components (\textbf{Left}) and prompt length (\textbf{Right}). We use 10-taks ImageNet-R setting.}
 }
 \label{fig:exp:prompt_component_abl}
\end{minipage}
\hspace{0.2cm}
\begin{minipage}[b]{0.33\linewidth}
\vspace{-1mm}
\centering
\includegraphics[width=0.99\textwidth]{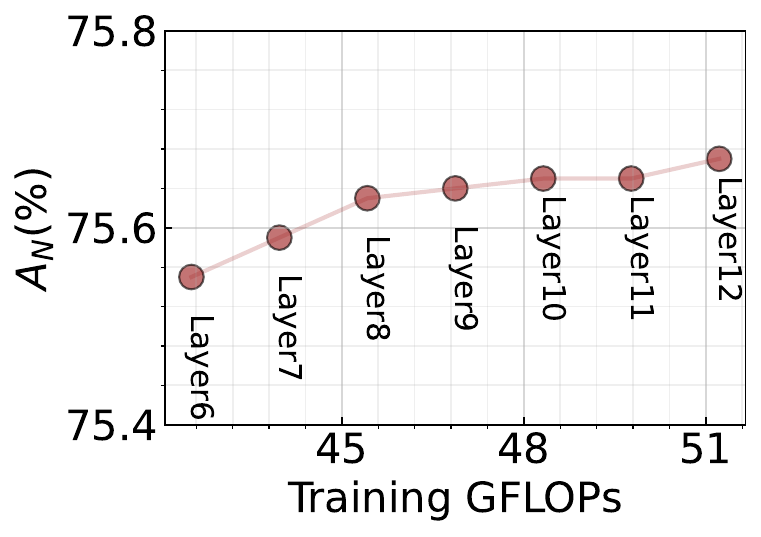}
        \vspace{-7mm}
 \caption{
Trade-off between GFLOPs and accuracy within a reference ViT. 
% We label each datapoint with layer index.
 }
\label{fig:exp:tradeoff_refvit}
\end{minipage}
\vspace{-9mm}
\end{figure}

\noindent\textbf{Prompt Design.} We further examine the impact of both the number of prompt components and the prompt length on accuracy. In Fig. \ref{fig:exp:prompt_component_abl} (left), we measure accuracy across configurations with $\{10,20,50,100,200,500\}$ prompts. The OS-prompt demonstrates a consistent accuracy enhancement as the number of prompts increases. Notably, OS-prompt++ reaches a performance plateau after just $50$ prompts. Additionally, our ablation study of prompt length, presented in Fig. \ref{fig:exp:prompt_component_abl} (right), reveals that our method maintains stable accuracy across various prompt lengths.

% \begin{wrapfigure}{h}{0.45\textwidth}
% % \begin{figure}[H]
%  \vspace{-5mm}
%      \centering
%          \includegraphics[width=0.41\textwidth]{figures/ref_layer_abl.pdf}
%         \vspace{-3mm}
%  \caption{
% Trade-off between GFLOPs and accuracy within a reference ViT. We label each datapoint with layer index.
%  }
%  \vspace{-5mm}
%      \label{fig:exp:tradeoff_refvit}
% % \end{figure}
% \end{wrapfigure}

\begin{table}[t]
 \aboverulesep=0.1ex % Solution part 1 of 3
   \belowrulesep=0ex % Solution part 1 of 3
\begin{minipage}[b]{0.48\linewidth}\centering
   \caption{ Analysis of QR loss design on 10-task ImageNet-R setting.
}
\vspace{-1mm}
\resizebox{0.99\textwidth}{!}{
\begin{tabular}{cc|cc}
% \hlinewd{1pt}
 \toprule
% Setting &  \multicolumn{2}{c}{Task-10} \\
%  \midrule
  CosSim & Softmax  & $A_N(\uparrow)$ & 
  $F_N(\downarrow)$ \\
  % CosSim & Softmax &  & \\
 \midrule
  &  &75.00 ± 0.53  & 1.68 ± 0.12\\
  & \checkmark &75.47 ± 0.42  & 1.38 ± 0.16\\
\checkmark& & 75.51 ± 0.33  & 1.28 ± 0.06\\
\checkmark & \checkmark& 75.67 ± 0.40 & 1.27 ± 0.10\\
% \hlinewd{1pt}
 \bottomrule
\end{tabular}
 \label{table:exp:qr_design}
}
\end{minipage}
\hspace{0.2cm}
\begin{minipage}[b]{0.499\linewidth}
\caption{Hyperparameter sensitivity study of $\lambda$ on ImageNet-R 5/10/20-task.
}
\vspace{-1mm}
\centering
\resizebox{0.99\textwidth}{!}{
\begin{tabular}{c|ccc}
% \hlinewd{1pt}
 \toprule
% Setting &  \multicolumn{2}{c}{Task-10} \\
%  \midrule
$\lambda$  & Task-5  & Task-10 & Task-20  \\
 \midrule
1e-5 &77.03 ± 0.10  & 75.63 ± 0.39 & 73.63 ± 0.21\\
5e-5 & 77.02 ± 0.13  & 75.62 ± 0.41 & 73.62 ± 0.19\\
1e-4 &  77.07 ± 0.15 & 75.67 ± 0.40 & 73.77 ± 0.19\\
5e-4 &  77.13 ± 0.24 & 75.68 ± 0.38 & 73.68 ± 0.17\\
% \hlinewd{1pt}
 \bottomrule
\end{tabular}
 \label{table:exp:hyperparameterabl}
}
\end{minipage}
\vspace{-1mm}
\end{table}

\begin{table}[t]
   \centering
% \small
 \aboverulesep=0.1ex % Solution part 1 of 3
   \belowrulesep=0ex % Solution part 1 of 3
   \caption{Analysis on prompt formation methods.  Results on ImageNet-R for different task configurations: 5 tasks (40 classes/task), 10 tasks (20 classes/task), and 20 tasks (10 classes/task). $A_N$ represents the average accuracy across tasks, and $F_N$ indicates the mean forgetting rate. We average values over five runs with different seeds.
}
\resizebox{0.85\textwidth}{!}{%
\begin{tabular}{l|cc|cc|cc}

% \hlinewd{1pt}
 \toprule
Setting &  \multicolumn{2}{c|}{Task-5} & \multicolumn{2}{c|}{Task-10} &\multicolumn{2}{c}{Task-20} \\
 \midrule
Method & $A_N(\uparrow)$ & $F_N(\downarrow)$ &  $A_N(\uparrow)$ & $F_N(\downarrow)$  &  $A_N(\uparrow)$ & $F_N(\downarrow)$  \\
 \midrule
UB & 77.13 & - & 77.13 & - & 77.13 & -\\
 \midrule
Deep L2P &73.93 ± 0.37& 2.69 ± 0.10 &71.66 ± 0.64 &1.78 ± 0.16 &68.42 ± 1.20 &1.12 ± 0.13  \\
OS-Prompt (L2P) & 74.61 ± 0.29 &2.29 ± 0.21 & 73.43 ± 0.60& 1.42 ± 0.10& 71.42 ± 0.24&
0.95 ± 0.06\\
% OS-Prompt++ (L2P) & &  &  &  &  & \\
 \midrule
 DualPrompt &73.05 ± 0.50& 2.64 ± 0.17 &71.32 ± 0.62& 1.71 ± 0.24& 67.87 ± 1.39& 1.07 ± 0.14\\
OS-Prompt (Dual) & 74.65 ± 0.15 &2.06 ± 0.20 & 72.57 ± 0.23& 1.13 ± 0.03& 70.55 ± 0.48&
0.81 ± 0.10\\
% OS-Prompt++ (Dual) & &  &  &  &  & \\
 \midrule
 CodaPrompt & 76.51 ± 0.38& 2.99 ± 0.19& 75.45 ± 0.56&1.64 ± 0.10 &72.37 ± 1.19& 0.96 ± 0.15 \\
OS-Prompt (Coda) & 75.74 ± 0.58 &3.32 ± 0.31 & 74.58 ± 0.56& 1.92 ± 0.15& 72.00 ± 0.60&
1.09 ± 0.11\\
% OS-Prompt++ (Coda) & {77.07 ± 0.15} & {2.23 ± 0.18} & {75.67 ± 0.40} & {1.27 ± 0.10} & {73.77 ± 0.19} & {0.79 ± 0.07}\\
% \hlinewd{1pt}
 \bottomrule
\end{tabular}%
}
\label{table:supply:matching_str}
  \vspace{-3mm}
\end{table}

\subsection{Accuracy-Efficiency Trade-off within a Reference ViT}
\vspace{-1mm}
During the training of OS-prompt++, our method does not improve the energy efficiency (Note, we achieve $\sim 50\%$ computational saving during inference). %observed in previous PCL methodologies. 
This arises from our approach of extracting the reference prompt query $r$ from the final layer of the reference ViT. To further enhance energy efficiency during the OS-prompt++ training, we delve into the trade-off between accuracy and efficiency within the Reference ViT. Instead of relying on the last layer's $[CLS]$ token embedding for the reference prompt query, we opt for intermediate token embeddings. Since our prompt pool is applied in layers $1$ to $5$ of the backbone ViT, we utilize the intermediate token embeddings from layers deeper than $5$ within the reference ViT. Fig. \ref{fig:exp:tradeoff_refvit} illustrates this trade-off with respect to the layer index where we get the reference prompt. Our findings show that utill layer 8, there is only a slight increase in accuracy, which suggests a potential to reduce GFLOPs without performance drop.

\subsection{Analysis on Prompt Formation Strategy}
\label{apdx:analysis_prompt_formation}

In our approach, the construction of prompts is achieved through a weighted summation of components within the prompt pool, mirroring the strategy employed by CodaPrompt. In this section, we delve into the impact of varying prompt formation strategies on performance. For comparative insight, we present results from two previous PCL methodologies: L2P and DualPrompt. While L2P selects 5 prompts from the internal prompt pool, DualPrompt distinguishes between general and task-specific prompts. Aside from the distinct prompt formation strategies, the configurations remain consistent.

In Table \ref{table:supply:matching_str}, we adopt the prompt formation strategies proposed in L2P and Dual, resulting in configurations denoted as OS-Prompt (L2P) and OS-Prompt (Dual). Our findings reveal the following:
(1) The effectiveness of OS-prompt varies based on the prompt formation technique employed. Notably, OS-Prompt (L2P) and OS-Prompt (Dual) yield lower accuracy than our original approach, which relies on CodaPrompt. This suggests that our method's performance could benefit from refined prompt formation techniques in future iterations.
(2) The OS-prompt framework, when integrated with L2P and Dual prompt formation strategies, outperforms the original Deep L2P and DualPrompt, a trend not observed with CodaPrompt. This may indicate that the prompt formation strategies of L2P and DualPrompt, which rely on hard matching (i.e., top-k), are more resilient than CodaPrompt's. Conversely, CodaPrompt employs soft matching, utilizing a weighted summation of all prompts based on proximity. This might be more susceptible to the diminished representation power from the intermediate layer's features. Simultaneously, a prompt query imbued with task-specific data (since our query token incorporates task prompts) appears to enhance representation within the prompt pool.

\section{Conclusion}
\vspace{-2mm}
In this paper, we introduce the OS-Prompt framework where we improve the efficiency of the conventional two-stage PCL in a simple yet effective manner. 
% By harnessing the power of intermediate token embeddings for prompts and introducing the Query-Pool Regularization (QR) loss, 
We save computational cost without performance drop on class-incremental continual learning scenarios. 
One of the potential limitations of our OS-Prompt++ is, in comparison to OS-Prompt (our light version), it introduces an increase in training computational cost from the reference ViT feedforward. This computational efficiency during training becomes important in the context of addressing the online continual learning problem \cite{ghunaim2023real} where the model needs rapid training on the given streaming data. 
To further improve the efficiency, we can adopt the early exit in \cite{phuong2019distillation}, where they make predictions in the middle layers.

\section*{Acknowledgement} 
This work was supported in part by CoCoSys, a JUMP2.0 center sponsored by DARPA and SRC, the National Science Foundation (CAREER Award, Grant \#2312366, Grant \#2318152), TII (Abu Dhabi), and the U.S. Department of Energy, Advanced Scientific Computing Research program, under the Scalable, Efficient and Accelerated Causal Reasoning Operators, Graphs and Spikes for Earth and Embedded Systems (SEA-CROGS) project (Project No. 80278). Pacific Northwest National Laboratory (PNNL) is a multi-program national laboratory operated for the U.S. Department of Energy (DOE) by Battelle Memorial Institute under Contract No. DE-AC05-76RL01830.

\bibliographystyle{splncs04}
\bibliography{main}

\begin{thebibliography}{10}
\providecommand{\url}[1]{\texttt{#1}}
\providecommand{\urlprefix}{URL }
\providecommand{\doi}[1]{https://doi.org/#1}

\bibitem{aljundi2018memory}
Aljundi, R., Babiloni, F., Elhoseiny, M., Rohrbach, M., Tuytelaars, T.: Memory aware synapses: Learning what (not) to forget. In: ECCV (2018)

\bibitem{buzzega2020dark}
Buzzega, P., Boschini, M., Porrello, A., Abati, D., Calderara, S.: Dark experience for general continual learning: a strong, simple baseline. In: NeurIPS (2020)

\bibitem{cai2020tinytl}
Cai, H., Gan, C., Zhu, L., Han, S.: Tinytl: Reduce memory, not parameters for efficient on-device learning. Advances in Neural Information Processing Systems  \textbf{33},  11285--11297 (2020)

\bibitem{caron2021emerging}
Caron, M., Touvron, H., Misra, I., J{\'e}gou, H., Mairal, J., Bojanowski, P., Joulin, A.: Emerging properties in self-supervised vision transformers. In: Proceedings of the IEEE/CVF international conference on computer vision. pp. 9650--9660 (2021)

\bibitem{chaudhry2018efficient}
Chaudhry, A., Ranzato, M., Rohrbach, M., Elhoseiny, M.: Efficient lifelong learning with a-gem. arXiv preprint arXiv:1812.00420  (2018)

\bibitem{chaudhry2019tiny}
Chaudhry, A., Rohrbach, M., Elhoseiny, M., Ajanthan, T., Dokania, P.K., Torr, P.H., Ranzato, M.: On tiny episodic memories in continual learning. arXiv preprint arXiv:1902.10486  (2019)

\bibitem{choi2021dual}
Choi, Y., El-Khamy, M., Lee, J.: Dual-teacher class-incremental learning with data-free generative replay. In: Proceedings of the IEEE/CVF Conference on Computer Vision and Pattern Recognition. pp. 3543--3552 (2021)

\bibitem{dosovitskiy2020image}
Dosovitskiy, A., Beyer, L., Kolesnikov, A., Weissenborn, D., Zhai, X., Unterthiner, T., Dehghani, M., Minderer, M., Heigold, G., Gelly, S., et~al.: An image is worth 16x16 words: Transformers for image recognition at scale. ICLR  (2021)

\bibitem{gao2022r}
Gao, Q., Zhao, C., Ghanem, B., Zhang, J.: R-dfcil: Relation-guided representation learning for data-free class incremental learning. In: European Conference on Computer Vision. pp. 423--439. Springer (2022)

\bibitem{ghunaim2023real}
Ghunaim, Y., Bibi, A., Alhamoud, K., Alfarra, M., Al~Kader~Hammoud, H.A., Prabhu, A., Torr, P.H., Ghanem, B.: Real-time evaluation in online continual learning: A new hope. In: Proceedings of the IEEE/CVF Conference on Computer Vision and Pattern Recognition. pp. 11888--11897 (2023)

\bibitem{hadsell2020embracing}
Hadsell, R., Rao, D., Rusu, A.A., Pascanu, R.: Embracing change: Continual learning in deep neural networks. Trends in cognitive sciences  (2020)

\bibitem{harun2023efficient}
Harun, M.Y., Gallardo, J., Hayes, T.L., Kanan, C.: How efficient are today's continual learning algorithms? In: Proceedings of the IEEE/CVF Conference on Computer Vision and Pattern Recognition. pp. 2430--2435 (2023)

\bibitem{hayes2019memory}
Hayes, T.L., Cahill, N.D., Kanan, C.: Memory efficient experience replay for streaming learning. In: ICRA (2019)

\bibitem{he2022parameter}
He, X., Li, C., Zhang, P., Yang, J., Wang, X.E.: Parameter-efficient fine-tuning for vision transformers. arXiv preprint arXiv:2203.16329  (2022)

\bibitem{hendrycks2020many}
Hendrycks, D., Basart, S., Mu, N., Kadavath, S., Wang, F., Dorundo, E., Desai, R., Zhu, T., Parajuli, S., Guo, M., Song, D., Steinhardt, J., Gilmer, J.: The many faces of robustness: A critical analysis of out-of-distribution generalization. arXiv preprint arXiv:2006.16241  (2020)

\bibitem{hendrycks2021many}
Hendrycks, D., Basart, S., Mu, N., Kadavath, S., Wang, F., Dorundo, E., Desai, R., Zhu, T., Parajuli, S., Guo, M., et~al.: The many faces of robustness: A critical analysis of out-of-distribution generalization. In: ICCV. pp. 8340--8349 (2021)

\bibitem{hsu2018re}
Hsu, Y.C., Liu, Y.C., Ramasamy, A., Kira, Z.: Re-evaluating continual learning scenarios: A categorization and case for strong baselines. arXiv preprint arXiv:1810.12488  (2018)

\bibitem{hu2021lora}
Hu, E.J., Shen, Y., Wallis, P., Allen-Zhu, Z., Li, Y., Wang, S., Wang, L., Chen, W.: Lora: Low-rank adaptation of large language models. arXiv preprint arXiv:2106.09685  (2021)

\bibitem{jia2022visual}
Jia, M., Tang, L., Chen, B.C., Cardie, C., Belongie, S., Hariharan, B., Lim, S.N.: Visual prompt tuning. In: Computer Vision--ECCV 2022: 17th European Conference, Tel Aviv, Israel, October 23--27, 2022, Proceedings, Part XXXIII. pp. 709--727. Springer (2022)

\bibitem{khattak2023maple}
Khattak, M.U., Rasheed, H., Maaz, M., Khan, S., Khan, F.S.: Maple: Multi-modal prompt learning. In: Proceedings of the IEEE/CVF Conference on Computer Vision and Pattern Recognition. pp. 19113--19122 (2023)

\bibitem{krizhevsky2009learning}
Krizhevsky, A., Hinton, G., et~al.: Learning multiple layers of features from tiny images  (2009)

\bibitem{li2019learn}
Li, X., Zhou, Y., Wu, T., Socher, R., Xiong, C.: Learn to grow: A continual structure learning framework for overcoming catastrophic forgetting. In: ICML. pp. 3925--3934. PMLR (2019)

\bibitem{li2017learning}
Li, Z., Hoiem, D.: Learning without forgetting. TPAMI  \textbf{40}(12),  2935--2947 (2017)

\bibitem{lopez2017gradient}
Lopez-Paz, D., Ranzato, M.: Gradient episodic memory for continual learning. NeurIPS  (2017)

\bibitem{mai2022online}
Mai, Z., Li, R., Jeong, J., Quispe, D., Kim, H., Sanner, S.: Online continual learning in image classification: An empirical survey. Neurocomputing  \textbf{469},  28--51 (2022)

\bibitem{pellegrini2021continual}
Pellegrini, L., Lomonaco, V., Graffieti, G., Maltoni, D.: Continual learning at the edge: Real-time training on smartphone devices. arXiv preprint arXiv:2105.13127  (2021)

\bibitem{peng2019moment}
Peng, X., Bai, Q., Xia, X., Huang, Z., Saenko, K., Wang, B.: Moment matching for multi-source domain adaptation. In: Proceedings of the IEEE/CVF international conference on computer vision. pp. 1406--1415 (2019)

\bibitem{phuong2019distillation}
Phuong, M., Lampert, C.H.: Distillation-based training for multi-exit architectures. In: Proceedings of the IEEE/CVF international conference on computer vision. pp. 1355--1364 (2019)

\bibitem{rebuffi2018efficient}
Rebuffi, S.A., Bilen, H., Vedaldi, A.: Efficient parametrization of multi-domain deep neural networks. In: Proceedings of the IEEE Conference on Computer Vision and Pattern Recognition. pp. 8119--8127 (2018)

\bibitem{rebuffi2017icarl}
Rebuffi, S.A., Kolesnikov, A., Sperl, G., Lampert, C.H.: icarl: Incremental classifier and representation learning. In: CVPR. pp. 2001--2010 (2017)

\bibitem{rusu2016progressive}
Rusu, A.A., Rabinowitz, N.C., Desjardins, G., Soyer, H., Kirkpatrick, J., Kavukcuoglu, K., Pascanu, R., Hadsell, R.: Progressive neural networks. arXiv preprint arXiv:1606.04671  (2016)

\bibitem{serra2018overcoming}
Serra, J., Suris, D., Miron, M., Karatzoglou, A.: Overcoming catastrophic forgetting with hard attention to the task. In: ICML. pp. 4548--4557 (2018)

\bibitem{smith2021always}
Smith, J., Hsu, Y.C., Balloch, J., Shen, Y., Jin, H., Kira, Z.: Always be dreaming: A new approach for data-free class-incremental learning. In: Proceedings of the IEEE/CVF International Conference on Computer Vision. pp. 9374--9384 (2021)

\bibitem{smith2023coda}
Smith, J.S., Karlinsky, L., Gutta, V., Cascante-Bonilla, P., Kim, D., Arbelle, A., Panda, R., Feris, R., Kira, Z.: Coda-prompt: Continual decomposed attention-based prompting for rehearsal-free continual learning. In: Proceedings of the IEEE/CVF Conference on Computer Vision and Pattern Recognition. pp. 11909--11919 (2023)

\bibitem{smith2023closer}
Smith, J.S., Tian, J., Halbe, S., Hsu, Y.C., Kira, Z.: A closer look at rehearsal-free continual learning. In: Proceedings of the IEEE/CVF Conference on Computer Vision and Pattern Recognition. pp. 2409--2419 (2023)

\bibitem{voigt2017eu}
Voigt, P., Von~dem Bussche, A.: The eu general data protection regulation (gdpr). A Practical Guide, 1st Ed., Cham: Springer International Publishing  \textbf{10}(3152676),  10--5555 (2017)

\bibitem{wang2022s}
Wang, Y., Huang, Z., Hong, X.: S-prompts learning with pre-trained transformers: An occam’s razor for domain incremental learning. Advances in Neural Information Processing Systems  \textbf{35},  5682--5695 (2022)

\bibitem{wang2022sparcl}
Wang, Z., Zhan, Z., Gong, Y., Yuan, G., Niu, W., Jian, T., Ren, B., Ioannidis, S., Wang, Y., Dy, J.: Sparcl: Sparse continual learning on the edge. Advances in Neural Information Processing Systems  \textbf{35},  20366--20380 (2022)

\bibitem{wang2022dualprompt}
Wang, Z., Zhang, Z., Ebrahimi, S., Sun, R., Zhang, H., Lee, C.Y., Ren, X., Su, G., Perot, V., Dy, J., et~al.: Dualprompt: Complementary prompting for rehearsal-free continual learning. In: European Conference on Computer Vision. pp. 631--648. Springer (2022)

\bibitem{wang2021learning}
Wang, Z., Zhang, Z., Lee, C.Y., Zhang, H., Sun, R., Ren, X., Su, G., Perot, V., Dy, J., Pfister, T.: Learning to prompt for continual learning. CVPR  (2022)

\bibitem{yin2020dreaming}
Yin, H., Molchanov, P., Alvarez, J.M., Li, Z., Mallya, A., Hoiem, D., Jha, N.K., Kautz, J.: Dreaming to distill: Data-free knowledge transfer via deepinversion. In: Proceedings of the IEEE/CVF Conference on Computer Vision and Pattern Recognition. pp. 8715--8724 (2020)

\bibitem{yoon2017lifelong}
Yoon, J., Yang, E., Lee, J., Hwang, S.J.: Lifelong learning with dynamically expandable networks. arXiv preprint arXiv:1708.01547  (2017)

\bibitem{zaken2021bitfit}
Zaken, E.B., Ravfogel, S., Goldberg, Y.: Bitfit: Simple parameter-efficient fine-tuning for transformer-based masked language-models. arXiv preprint arXiv:2106.10199  (2021)

\bibitem{zenke2017continual}
Zenke, F., Poole, B., Ganguli, S.: Continual learning through synaptic intelligence. In: ICML (2017)

\bibitem{zhang2020side}
Zhang, J.O., Sax, A., Zamir, A., Guibas, L., Malik, J.: Side-tuning: a baseline for network adaptation via additive side networks. In: Computer Vision--ECCV 2020: 16th European Conference, Glasgow, UK, August 23--28, 2020, Proceedings, Part III 16. pp. 698--714. Springer (2020)

\bibitem{zhang2021tip}
Zhang, R., Fang, R., Zhang, W., Gao, P., Li, K., Dai, J., Qiao, Y., Li, H.: Tip-adapter: Training-free clip-adapter for better vision-language modeling. arXiv preprint arXiv:2111.03930  (2021)

\bibitem{zhou2022learning}
Zhou, K., Yang, J., Loy, C.C., Liu, Z.: Learning to prompt for vision-language models. International Journal of Computer Vision  \textbf{130}(9),  2337--2348 (2022)

\end{thebibliography}
\end{document}